\documentclass[sigconf,nonacm]{acmart}

\usepackage{booktabs}
\usepackage{amsmath}
\usepackage{graphicx}

\newcommand{\method}{STEP-D}
\newcommand{\eu}{\textsc{eu}}
\newcommand{\grpo}{\textsc{grpo}}

\begin{document}

\title{From Prediction to Decision: Decision-Focused Fine-Tuning of Autoregressive Tabular Transformers for Cost-Sensitive Financial Events}

\author{Ashkan Golgoon}
\orcid{0000-0002-0251-4203}
\affiliation{%
  \institution{AI Center of Excellence, U.S. Bank}
  \country{United States}
}
\email{ashkan.golgoon@usbank.com}

\author{Giacomo Domeniconi}
\orcid{0000-0003-2797-1547}
\affiliation{%
  \institution{AI Center of Excellence, U.S. Bank}
  \country{United States}
}
\email{giacomo.domeniconi@usbank.com}

\author{Samuel Atkins}
\orcid{0009-0004-2909-3180}
\affiliation{%
  \institution{AI Center of Excellence, U.S. Bank}
  \country{United States}
}
\email{samuel.atkins@usbank.com}

\author{Sergio Servantez}
\orcid{0000-0002-8294-0279}
\affiliation{%
  \institution{AI Center of Excellence, U.S. Bank}
  \country{United States}
}
\email{sergio.servantez@usbank.com}

\author{Zain Khandwala}
\orcid{0009-0009-1394-8814}
\affiliation{%
  \institution{AI Center of Excellence, U.S. Bank}
  \country{United States}
}
\email{zain.khandwala1@usbank.com}

\begin{abstract}
Autoregressive transformers trained on tokenized tabular event streams are a simple and strong baseline for predicting financial events such as fraud. Yet institutions do not act on probabilities; they act on decisions whose consequences are asymmetric and heteroscedastic: the cost of a missed high-value fraud and that of a wrongly declined small purchase are not the same. The standard bridge is post-hoc Bayes-optimal thresholding of calibrated probabilities under a cost matrix. We ask whether fine-tuning the model to the decision objective can do better. We fine-tune the same pretrained model so that its restricted softmax over label tokens (already a distribution over business actions) maximizes expected utility under an institution-specified, amount-scaled cost matrix. Because the reward is a known function of (action, outcome, amount) and the outcome is observed in the data, the expected-utility objective and its gradient are available in closed form. This is example-dependent cost-sensitive learning whose per-example loss is linear in the predicted probability, with no sampling, critics, or advantage estimation. A KL anchor to the pretrained model and an auxiliary language-modeling term preserve calibration and the generative event model. A sampled, group-normalized (GRPO-style) variant provides the extension point to trajectory-level rewards. On a high-imbalance card-fraud task with genuine decision headroom ($0.11\%$ prevalence, amount-scaled stakes), decision-focused fine-tuning improves realized utility per $1{,}000$ decisions over the Bayes-thresholded cross-entropy model ($+0.08$; $70\%\!\to\!76\%$ of the oracle gap closed, $\approx\!{+}\$12$ per $1{,}000$ screened transactions), consistently across all ten seeds (paired two-sided Wilcoxon $p{=}0.002$), and dramatically outperforms naive cost-sensitive training heuristics (class weighting, focal loss, oversampling), which collapse to over-intervention under extreme imbalance. Granting those heuristics the same KL anchor recovers only a fifth of the gap, so the gain is attributable to the objective's form rather than to regularization alone. Rolling out held-out future transaction windows, we further show that the anchors preserve STEP's generative event model (the fine-tuned model reproduces the fraud-label marginal and amount distribution of real continuations), whereas the same heuristics distort the generated fraud rate by $9$--$136\times$. The method requires no architectural changes and no data resampling, and adds a small fraction of pretraining compute.
\end{abstract}



\maketitle

\section{Introduction}
Event-level prediction problems are ubiquitous in finance: will this card transaction turn out to be fraudulent, will this client churn next quarter, will this loan default? A recent line of work shows that a deliberately simple recipe---tokenize each field of each tabular event, concatenate events into a sequence, and train a GPT-style decoder from scratch with next-token prediction---matches or beats specialized tabular architectures on exactly these tasks~\cite{step,padhi2021tabbert,zhang2023fatatrans}. The appeal is operational as much as scientific: one generic model, one generic objective, no feature engineering.

But cross-entropy answers the wrong question. Production systems built on these predictors must decide: block or pass the transaction, more generally, take or withhold a costly action. The losses attached to those decisions are (i) wildly asymmetric across error types, (ii) heteroscedastic, i.e., they scale with the amount or exposure of the specific event, and (iii) institution-specific. The standard answer is post hoc: calibrate the model's probabilities and threshold them at the Bayes-optimal point implied by the cost matrix~\cite{elkan2001foundations}. This is a strong baseline that we implement and report. The natural question is whether fine-tuning the model to the decision objective can beat this post-hoc rule. Under the extreme class imbalance typical of fraud ($\approx$0.1\% positives), cross-entropy spends almost all of its gradient signal on easy negatives, and the usual countermeasures such as oversampling, class weighting, focal loss~\cite{lin2017focal}, distort calibration and require dataset-specific tuning; we show they can fail badly.

We optimize the deployed quantity directly. Our starting point is that an autoregressive tabular transformer \emph{already contains a decision layer}: at every position where the next token belongs to the label column, the model's softmax restricted to the $K$ label-value tokens is a distribution over the $K$ possible business actions. We therefore fine-tune the pretrained model so that this restricted softmax maximizes expected utility under the institution's cost matrix (affine in the event's amount), with the pretrained model itself frozen as a reference that anchors the policy through a KL penalty, in the spirit of RLHF~\cite{ouyang2022rlhf}.

Crucially, this is not a generic RLHF pipeline transplanted onto tables, and it is more accurately described as decision-focused, example-dependent cost-sensitive learning~\cite{bahnsen2014example,elmachtoub2022spo} than as reinforcement learning. Because the reward is a known function of (action, outcome, amount) and the outcome is in the training data, the payoff of every counterfactual action is available. The decision problem is a contextual bandit with full feedback, so the expected-utility objective and its gradient are available in closed form: exact, zero-variance, and requiring no sampling, critics, or advantage estimation. As we show (\S\ref{sec:linear}), the resulting per-example loss is linear in the predicted probability, i.e., an amount-weighted cost-sensitive classification loss plus a generative anchor. We retain a sampled, group-normalized REINFORCE variant in the style of GRPO~\cite{shao2024grpo} as the extension point to multi-step rewards (utilities over generated future trajectories), where exact expectations are unavailable.

\paragraph{Contributions.}
\begin{enumerate}
\item \textbf{A decision-theoretic reframing} of autoregressive tabular event models: the restricted label softmax is a decision layer, and event prediction becomes example-dependent cost-sensitive decision-making with amount-scaled stakes.
\item \textbf{A closed-form expected-utility fine-tuning objective} for matrix-specified financial rewards, which we show is linear in the predicted probability (\S\ref{sec:linear}), with KL and language-modeling anchors that recover the pretrained model in the large-anchor limit; plus a GRPO-style sampled counterpart for rollout-based rewards.
\item \textbf{A utility-denominated evaluation protocol} for event transformers---realized utility per 1{,}000 decisions against Bayes-threshold, always-pass, and oracle references, with cost curves and cost-misspecification sweeps---which we argue should accompany AUC in this literature.
\item \textbf{A focused empirical study} on a public high-imbalance card-fraud dataset showing (a) a small but consistent utility gain over post-hoc thresholding, and a large gain over cost-sensitive training heuristics; and (b) a generative-fidelity analysis (rolling out future event windows) confirming that the KL/LM anchors preserve the pretrained event model (reproducing the real fraud-label marginal and amount distribution) while the same cost-sensitive heuristics that damage calibration also corrupt the generative model.
\end{enumerate}

\section{Related Work}
\paragraph{Transformers for tabular event streams.}
TabBERT~\cite{padhi2021tabbert} and FATA-Trans~\cite{zhang2023fatatrans} encode transaction sequences with BERT-style objectives and field-aware embeddings; STEP~\cite{step} shows that a plain decoder-only model over word-level ``column\_value'' tokens, trained from scratch with next-token prediction, is competitive while being simpler. Event Stream GPTs~\cite{mcdermott2023esgpt} and neural temporal point processes such as Transformer Hawkes Process~\cite{zuo2020thp} generatively model sequences of marked events in continuous times. All of these optimize likelihoods and report discrimination metrics; none optimizes, or evaluates in, the currency of the downstream decision. Our method is a post-pretraining stage agnostic to which sequence model is underneath; we instantiate it on STEP because of its simplicity and strong baseline status.

\paragraph{Cost-sensitive and imbalanced learning.}
Classical remedies include class-weighted losses, resampling, focal loss~\cite{lin2017focal}, and post-hoc threshold moving; the Bayes decision rule for a known cost matrix is textbook decision theory~\cite{elkan2001foundations}. A long line of work makes an arbitrary base learner cost-sensitive without touching its loss, by relabelling training data under the estimated cost-minimizing class (MetaCost~\cite{domingos1999metacost}), by reweighting boosting rounds with misclassification costs (AdaCost~\cite{fan1999adacost}, CSB~\cite{ting2000csb}), or by cost-proportionate example weighting with rejection sampling~\cite{zadrozny2003costproportionate}. These are the intellectual ancestors of our B5--B8 baselines, and they share the property we find decisive here: they alter the \emph{training distribution or loss}, and therefore, the probability estimates, which is exactly what the Bayes decision rule can least afford under $0.1\%$ prevalence. Example-dependent cost-sensitive learning, where, e.g., the cost of a missed fraud is the transaction amount, has been studied for shallow classifiers~\cite{bahnsen2014example}. We bring example-dependent costs to pretrained autoregressive event models. There the classifier is one slice of a generative model, and naively changing the loss would distort both its probability estimates and the rest of the model; we therefore anchor training with a KL/LM term, which has no analogue in the shallow setting.
%
%
Post-hoc recalibration~\cite{guo2017calibration} is the complementary move, and we include it as B4.

\paragraph{RL fine-tuning of pretrained sequence models.}
RLHF~\cite{ouyang2022rlhf} and GRPO~\cite{shao2024grpo} align language models to reward signals while regularizing toward the pretrained policy with a KL term. We adopt the same architecture of objective but exploit the structure of our rewards: full counterfactual feedback makes the gradient exact, removing the estimator-variance machinery (critics, group baselines) those methods exist to manage. Our sampled \grpo{} variant is thus a controlled ablation: same objective, with and without sampling noise.

\paragraph{Decision-focused learning.}
A parallel literature trains predictive models through the downstream optimization problem, either by differentiating the optimizer~\cite{donti2017task,wilder2019melding} or by surrogate regret losses such as SPO+~\cite{elmachtoub2022spo}; see~\cite{mandi2024dfl} for a survey. Closest to ours are methods that apply decision-focused learning as a second stage on top of an already-trained predictor, rather than training end-to-end from scratch. Examples include decision-focused retraining of forecasters in smart energy systems~\cite{beichter2024decision}, and DFF~\cite{yang2025dff}, which fine-tunes a predict-then-optimize pipeline under limited data. We share that two-stage structure (a cheap decision-aware stage appended to a model pretrained for accuracy) and differ in two ways. First, those settings face a combinatorial or continuous optimization layer and must contend with non-differentiable or degenerate argmins. Our per-event discrete action is simple enough that the decision layer is differentiable in closed form, which is precisely what yields the exact gradient and removes the surrogate entirely. Second, and more consequentially, the predictor being retrained there is a \emph{discriminative} forecaster, whereas ours is a \emph{generative} event model whose other columns must survive the procedure---which is why anchoring is required and why we evaluate generative fidelity (\S\ref{sec:genfid}). This work is the event-sequence instantiation of decision-focused learning.

\paragraph{Background: Simple Transformer for Event Prediction}
STEP~\cite{step} tokenizes each row of an event table into per-column word-level tokens ``loc\_value'' (numeric columns quantile-bucketed into 32 bins; time encoded as time since last event), concatenates $L$ consecutive rows of one user into a sequence framed by BOS/EOS, and trains a GPT-2-shaped decoder from scratch with causal cross-entropy over all tokens. During training, label-column tokens in the input are stochastically replaced by a mask token so the model cannot copy the answer; at inference the label slot is masked and the model's next-token distribution at that position, restricted to the $K$ label-value tokens and renormalized,
\begin{equation}
p_\theta(y=k \mid h) \;=\; \frac{\exp z_\theta(\tau_k \mid h)}{\sum_{j=1}^{K}\exp z_\theta(\tau_j \mid h)},
\label{eq:restricted}
\end{equation}
where $\tau_k$ is the token of label value $k$, is used for classification. We change nothing about this stage.

\section{Method: Reward-Aligned Fine-Tuning}
\paragraph{The policy hiding in the pretrained model}
Fix a decision task with $K$ actions identified with the $K$ label values; in our task $K{=}2$, with $a{=}1$ the \emph{intervention} (block the transaction) and $a{=}0$ \emph{pass}. At every sequence position $t$ whose next token is the label slot of event row $r$, define the policy
\begin{equation}
\pi_\theta(a \mid h_t) \;:=\; p_\theta(y=a \mid h_t)
\end{equation}
via Eq.~\eqref{eq:restricted}. Pretraining shapes $\pi$ toward the conditional label distribution; fine-tuning reshapes it toward the utility-maximizing action distribution.

\paragraph{Amount-dependent utility}
The institution specifies a cost matrix whose entries are affine in the event's monetary stake $m_r$ (the amount column of the same row, decoded from its quantile bucket to a representative value):
\begin{equation}
R(a,y,m)\in\{r_{tp},r_{fp},r_{fn},r_{tn}\},\qquad r_{\bullet}= \phi_{\bullet} + \gamma_{\bullet}\, m .
\end{equation}
Section~\ref{sec:setup} gives the concrete matrix with its business justification. Under a calibrated probability $p$, the Bayes-optimal rule is to intervene iff $p \ge t^*(m)$ with
\begin{equation}
t^*(m) \;=\; \frac{r_{tn}-r_{fp}}{(r_{tn}-r_{fp}) + (r_{tp}-r_{fn})},
\label{eq:bayes}
\end{equation}
which is amount-dependent whenever any $\gamma_\bullet \neq 0$. This rule applied to the \emph{pretrained} model's probabilities is our strongest baseline.

\paragraph{Closed-form expected-utility objective}
Because $y_r$ is observed and $R$ is known, the reward of every action is known---full counterfactual feedback. The expected utility at a decision point,
\begin{equation}
J_t(\theta) \;=\; \sum_{a} \pi_\theta(a\mid h_t)\, R(a, y_r, m_r),
\end{equation}
is an explicit differentiable function of the logits; no sampling, score-function estimator, baseline, or critic is needed. Our Stage-2 loss over a batch $\mathcal{B}$ of decision points is
\begin{equation}
\mathcal{L}(\theta) = \frac{1}{|\mathcal{B}|}\!\sum_{t\in\mathcal{B}}\!\Big[-\tfrac{J_t(\theta)}{s} + \beta\, \mathrm{KL}\!\big(\pi_\theta(\cdot|h_t)\,\|\,\pi_{\mathrm{ref}}(\cdot|h_t)\big)\Big] + \lambda\, \mathcal{L}_{\mathrm{CE}}(\theta),
\label{eq:loss}
\end{equation}
where $s$ is a fixed reward scale, $\pi_{\mathrm{ref}}$ is the frozen pretrained model evaluated under the \emph{identical} stochastic label-masking pattern (we replay the RNG state), and $\mathcal{L}_{\mathrm{CE}}$ is the ordinary causal cross-entropy over all tokens. $\beta$ controls how far the label policy may drift from the calibrated pretrained distribution; $\lambda$ prevents degradation of the non-label columns, preserving the generative event model. We call this objective \eu.

\eu{} is the exact expectation of REINFORCE for this one-step bandit (Appendix~A): zero gradient variance, one forward--backward per batch plus a no-grad reference forward. It addresses class imbalance \emph{without resampling}: the gradient magnitude at each decision point is proportional to the utilities at stake, so a rare high-value fraud contributes commensurately more than an abundant low-value legitimate purchase.

\paragraph{The objective is linear in the predicted probability}
\label{sec:linear}
For the binary case ($K{=}2$, $a\in\{0,1\}$), writing $p=\pi_\theta(1\mid h_t)$ and using $R(a,y,m)$ from the cost matrix,
\begin{equation}
J_t(\theta) = (1{-}p)\,R(0,y_r,m_r) + p\,R(1,y_r,m_r) = R(0,y_r,m_r) + p\,\Delta_t,
\end{equation}
with \emph{example-dependent} margin $\Delta_t = R(1,y_r,m_r)-R(0,y_r,m_r)$. The utility term of Eq.~\eqref{eq:loss} is therefore \emph{linear in $p$}: $-J_t/s$ is an amount-weighted cost-sensitive classification loss with per-example weight $|\Delta_t|$ and target $\mathbf{1}[\Delta_t{>}0]$. This exposes what the method is (example-dependent cost-sensitive learning~\cite{bahnsen2014example}, not generic RL) and what the anchors are for: unregularized, a loss linear in $p$ is minimized at the vertices $p\in\{0,1\}$ (degenerate, uncalibrated); the KL term $\beta\,\mathrm{KL}(\pi_\theta\|\pi_{\mathrm{ref}})$ is the load-bearing regularizer that keeps $p$ near the calibrated pretrained probability while the margin tilts it toward the higher-utility action. $\beta$ is thus not a minor hyperparameter but the term that separates useful reallocation from collapse.

\paragraph{GRPO-style sampled variant}
\label{sec:grpo}
For rewards requiring rollouts of future events the expectation is intractable. To quantify what sampling costs---and to provide the extension point---we implement group-sampled REINFORCE: at each decision point draw $a_1{,}\dots{,}a_G \sim \pi_\theta$, form within-group advantages $\hat A_i$, and ascend $\sum_i \hat A_i \log \pi_\theta(a_i)$ with the same KL/LM anchors. For the advantage we use a leave-one-out (RLOO) baseline $\hat A_i = R_i - \tfrac{1}{G-1}\sum_{j\neq i} R_j$; because a baseline is a control variate, its expected gradient is \emph{exactly} the \eu{} gradient (Appendix~A), and this is the variant we report as \grpo. The canonical std-normalized advantage $\hat A_i = (R_i - \bar R)/(\sigma_R + \varepsilon)$ instead divides by a within-group reward standard deviation that vanishes whenever a peaked policy makes all $G$ samples agree; under the extreme class imbalance here this can collapse the intervention rate and destroy utility, which is precisely why we adopt the RLOO baseline. A single gradient step per batch keeps the procedure on-policy (no ratio clipping needed).

\paragraph{Decision scope and anchors}
Decisions are trained at \emph{all} label slots (every event row; maximal signal; natural for transaction-level fraud). As $\beta\!\to\!\infty$, $\pi_\theta\!\to\!\pi_{\mathrm{ref}}$ and Stage-1 behavior is recovered; as $\lambda\!\to\!\infty$ all non-label predictions are pinned. At finite $(\beta,\lambda)$ the method interpolates between the pretrained event model and the pure utility maximizer; we use fixed $(\beta,\lambda)$ throughout and leave a full grid search to future work.

\section{Experimental Setup}
\label{sec:setup}

\paragraph{Dataset and decision task}
The dataset is public and was used in the original STEP study (so Stage-1 numbers are directly comparable), and instantiates a high-frequency financial screening archetype: card-fraud detection.

\paragraph{D1 --- Synthetic card transactions (IBM), fraud screening.}
24.4M card transactions from 2{,}000 simulated consumers. Fraud prevalence is $0.122\%$ over the full corpus and $0.112\%$ on the held-out users we evaluate on; we quote the latter ($\approx0.11\%$) throughout, since it is the rate the decision policy actually faces.\footnote{Per-split rates: train $0.121\%$, validation $0.140\%$, test $0.112\%$. The spread is sampling noise across the 200-user test draw, not a distribution shift.} Decision: \emph{block (review) vs.\ pass} each transaction, made at every event (decision scope \emph{all}).

\emph{Stakes.} The amount column is quantile-bucketed into 32 equal-frequency bins by recursive median splitting, as in STEP, and the reward uses a monotone function $m$ of the bucket as a severity proxy for the transaction's stake rather than raw dollars. Bucket medians run from $\$1.21$ to $\$251.43$ (overall median ticket $\$30$). Because the choice of severity scale is a modelling decision and not a result, we re-score every method under three scales in \S\ref{sec:scale_robustness}: the scale the models were trained under, the normalized bucket rank $(b{+}0.5)/32$, and a dollar-proportional scale given by the bucket medians. The ranking of methods, the sign and significance of every comparison, and the gap-closed percentages are unchanged across all three; the dollar scale additionally lets us state the headline gain in money.

\emph{Refunds.} $5.10\%$ of rows carry a negative \texttt{Amount} (refunds and reversals), including $1{,}125$ fraud rows ($3.78\%$ of all fraud, $8.20\%$ of fraud by $|\$|$). Under quantile bucketing these fall into the two lowest bins and are therefore treated as the lowest-stake events, even though a large refund is a large exposure. This under-weights a real fraud mode; we do not correct it, because doing so would change the task definition relative to prior work on this corpus, but it means our reported utilities are a conservative lower bound on the achievable gain and that $|\texttt{Amount}|$ would be the better severity input in a deployment.

\paragraph{Cost matrix}
\label{sec:costs}
Table~\ref{tab:costs} gives the primary cost matrix ($r = \phi + \gamma m$); Section~\ref{sec:robustness_results} sweeps it to show conclusions are not an artifact of one choice. Magnitudes follow practitioner conventions: interchange revenue $\approx$20\,bps of ticket. The implied Bayes threshold $t^*(m)$ (Eq.~\ref{eq:bayes}) is reported with the results; the threshold is amount-dependent, which already makes the thresholding baseline nontrivial.

\begin{table}[t]
\caption{Primary cost matrix. Entries are $\phi + \gamma m$ with $m$ the row's stake. $m$ is a monotone severity proxy derived from the amount bucket, normalized to $(0,1]$, not a dollar figure (\S\ref{sec:setup}); \S\ref{sec:scale_robustness} re-scores every result under two alternative scalings of $m$, including a dollar-proportional one.}
\label{tab:costs}
\begin{tabular}{@{}lc@{}}
\toprule
 & D1 fraud \\
\midrule
$r_{tp}$ (intervene, $y{=}1$) & $1.00\,m - 0.02$ \\
$r_{fp}$ (intervene, $y{=}0$) & $-0.05 - 0.05\,m$ \\
$r_{fn}$ (pass, $y{=}1$)      & $-1.00\,m$        \\
$r_{tn}$ (pass, $y{=}0$)      & $+0.002\,m$       \\
\bottomrule
\end{tabular}
\end{table}

\paragraph{Models and training}
Stage 1 reuses the STEP configuration (GPT-2-small-shaped decoder trained from scratch; word-level vocabulary $\le$60k; sequence length 10 rows; stochastic label masking). Stage-1 hyperparameters are tuned on a proper validation split (Sec.~\ref{sec:protocol}), never on test.

Stage 2 fine-tunes the Stage-1 checkpoint with \eu{} or \grpo: AdamW, lr $10^{-6}$, $\beta{=}0.05$, $\lambda{=}0.1$, reward scale $s = 1.0$, group size $G{=}8$. We run 3 epochs over a fixed $10^5$-sequence subsample ($\sim$1M decisions) to keep Stage 2 a small fraction of Stage-1 cost. Every cost-sensitive baseline below is granted the \emph{same} Stage-2 budget on the same Stage-1 checkpoint, so all comparisons are matched-compute continuations (App.~B). The anchor coefficients are swept in \S\ref{sec:beta_sweep}.

\paragraph{Splits and protocol}
\label{sec:protocol}
D1 is split \emph{by client}: 1{,}600 training users, 200 validation users, 200 test users, drawn disjointly, so no client appears in two splits and no user's history is ever split across the train/test boundary. All three splits span the full 1991--2020 period; the split is not temporal.\footnote{This departs from the temporal (train $<$2018) convention used in the original STEP study. The user-level split is the appropriate one here because our unit of decision is a client's event stream and we need whole, uncut histories to condition on, but it means our results measure generalization to \emph{new clients}, not robustness to \emph{temporal drift}. We do not claim the latter; a rolling-origin evaluation is the natural next test and we flag it in \S\ref{sec:limitations}.} All model selection---Stage-2 epochs, $(\beta,\lambda)$, and baseline thresholds---uses \textbf{validation utility} only. Test is touched once per (method $\times$ seed). Ten seeds; we report mean $\pm$ s.d.\ and a paired two-sided Wilcoxon test across seeds for the headline comparison (\eu{} vs.\ CE+Bayes).

Within a seed, every STEP-family method scores the \emph{identical} decision set (same users, same windows, same label masking), so all comparisons are exactly paired; we assert this invariant programmatically rather than assuming it. One consequence of the inherited stochastic label masking is that $10.0\%$ of decisions have their amount token masked and hence an unreadable stake. Under the cost matrix of Table~\ref{tab:costs} a zero stake implies $t^*{=}1$, so those events are passed by every Bayes-thresholded method and contribute exactly zero utility; they are inert and identical across arms, but they do dilute the per-1{,}000 denominator by a tenth. Utilities should therefore be read as per 1{,}000 \emph{scored} events, of which $90\%$ carry a live stake.

\paragraph{Baselines}
\label{sec:baselines}
Grouped by the question they answer.

\emph{Does sequence pretraining matter at all?}
\textbf{B1} logistic regression and \textbf{B2} gradient-boosted trees (HistGBM) on standard engineered features (per-user aggregates; recency/frequency/monetary; current-event fields) with class weights and Bayes thresholding---the ``what a bank actually runs'' reference.

\emph{Is post-hoc decision theory enough?}
\textbf{B3} \textbf{STEP-CE + Bayes threshold} $t^*(m)$---the strongest honest baseline and our primary comparator. \textbf{B4} B3 with post-hoc recalibration fit on validation and applied to test, in three increasingly expressive forms: \textbf{B4a} temperature scaling (one parameter), \textbf{B4b} Platt scaling~\cite{platt1999probabilistic} (slope and intercept), and \textbf{B4c} isotonic regression~\cite{zadrozny2002isotonic} (nonparametric). B4c matters disproportionately: \emph{every} post-hoc calibrator is a monotone map of the predicted scores, and isotonic regression is the most expressive monotone map there is. It is therefore not merely another baseline but an empirical \emph{upper bound on the entire class}---if it does not close the gap to \method{}, no recalibration scheme can. We verify monotonicity of each fitted map directly rather than inferring it.

\emph{Do cost-sensitive training heuristics do better?} As matched-compute Stage-2 continuations of the same Stage-1 checkpoint: \textbf{B5} class-weighted cross-entropy (inverse-frequency positive weight), \textbf{B6} focal loss~\cite{lin2017focal}, and \textbf{B7} random oversampling of positives. These are the standard imbalance remedies and the sharpest test of whether the closed-form utility objective is needed.

\emph{Is the KL anchor doing the work, or the loss form?} \textbf{B8} \textbf{class-weighted CE + KL anchor}: B5 with the \emph{same} KL-to-reference term and coefficient $\beta{=}0.05$ that \method{} uses, and the same Stage-2 budget. B5--B7 as usually specified carry no anchor, so a comparison against them alone confounds two changes---the objective's form and the presence of a calibration anchor. B8 breaks that confound: B8 vs.\ B5 isolates the anchor's contribution, and M1 vs.\ B8 isolates the contribution of the expected-utility loss form once both are anchored. This is the decisive ablation for our claim and we report it as a first-class row.

\emph{Ours.} \textbf{M1} \method{} (\eu) and \textbf{M2} \method{} (\grpo), decisions via Bayes-thresholding of the fine-tuned probabilities.

\emph{Context lines.} Always-pass and the oracle (intervene iff $y{=}1$) bound the achievable range and normalize utility as ``\% of oracle gap closed.''

\paragraph{Metrics}
Primary: \textbf{realized utility per 1{,}000 decisions} on test, with its decomposition into TP/FP/FN/TN contributions. Discrimination: PR-AUC (key under imbalance), ROC-AUC, F1. Calibration: ECE (15 bins). Operational: intervention rate; precision/recall at the operating point. All methods in the STEP family score the \emph{identical} decision set per seed (shared oracle and always-pass floors), enforced by assertion in the aggregation script.

\paragraph{Experimental Questions}
\label{sec:plan}
Our study is organized around three questions, answered in \S\ref{sec:results_d1}--\S\ref{sec:robustness_results}. \textbf{(Q1)} On a high-imbalance task with amount-scaled stakes and genuine decision headroom (D1 fraud), does closed-form utility fine-tuning improve realized utility over Bayes-thresholded cross-entropy (B3/B4), and does it avoid the calibration damage of cost-sensitive training heuristics (B5--B7)? \textbf{(Q2)} Is any gain attributable to the \emph{objective}, or merely to the KL anchor that \method{} happens to carry and the heuristics do not (B8)? \textbf{(Q3)} Are the D1 gains robust to misspecification of the false-positive cost (\S\ref{sec:robustness_results}) and to the choice of severity scale (\S\ref{sec:scale_robustness})? We use ten seeds, with a pre-registered paired two-sided Wilcoxon test for the headline comparison (\eu{} vs.\ B3).

\begin{table*}[t]
\caption{Main results (mean $\pm$ s.d.\ over 10 seeds). Utility per 1{,}000 decisions; \%gap = share of the always-pass$\to$oracle gap closed. STEP methods (B3--M2) score the \emph{identical} decision set per seed (shared oracle/floors), enforced by assertion. \textbf{B1/B2 are computed on a separate user-level harness} (one decision per client, higher positive rate) with a \emph{different oracle}, shown in brackets $[o]$ after their utility; their \%gap is relative to their own oracle. B1/B2 utility/1k is therefore \emph{not comparable} to the STEP rows. ($^{\ddagger}$marks the separate-harness rows.) Best utility among matched STEP methods in \textbf{bold}; the best ECE is bolded separately. The Focal (B6) \%gap is omitted: with utility far below the always-pass floor the normalized gap is uninformative. B8 is B5 plus \method{}'s KL anchor, and separates the anchor from the objective (\S\ref{sec:anchor_vs_loss}).}
\label{tab:main}
\begin{tabular}{@{}lcccccc@{}}
\toprule
Method & Utility/1k & \%gap & PR-AUC & ROC-AUC & ECE & Interv.\ rate \\
\midrule
Always-pass                & $0.19 \pm 0.06$   & $0$     & --    & --    & --     & $0$ \\
B1 LogReg + Bayes$^{\ddagger}$  & $-0.09 \pm 0.03\;[o\,0.66]$  & $0.5$   & 0.014 & 0.760 & 0.0001 & 0.0002 \\
B2 HistGBM + Bayes$^{\ddagger}$ & $-0.00 \pm 0.06\;[o\,0.66]$  & $12.4$  & 0.098 & 0.842 & 0.0003 & 0.0010 \\
\cmidrule(l){1-7}
B3 STEP-CE + Bayes         & $1.173 \pm 0.069$ & $70.4$  & 0.690 & 0.972 & 0.00030 & 0.0012 \\
B4a \;+ TempScale         & $1.183 \pm 0.062$ & $71.1$  & 0.690 & 0.972 & 0.00020 & 0.0012 \\
B4b \;+ Platt             & $1.199 \pm 0.065$ & $72.2$  & 0.690 & 0.972 & 0.00010 & 0.0014 \\
B4c \;+ Isotonic          & $1.204 \pm 0.065$ & $72.6$  & 0.679 & 0.972 & \textbf{0.00008} & 0.0015 \\
B5 Weighted-CE            & $-0.76 \pm 0.73$  & $-69$   & 0.695 & 0.985 & 0.0126 & 0.0244 \\
B6 Focal                  & $-28.8 \pm 6.4$   & --      & 0.693 & 0.983 & 0.0948 & 0.3122 \\
B7 Oversample             & $0.06 \pm 0.59$   & $-10$   & 0.724 & 0.984 & 0.0083 & 0.0157 \\
B8 Weighted-CE + KL       & $-0.34 \pm 0.64$  & $-39$   & 0.697 & 0.984 & 0.0107 & 0.0199 \\
M1 \method{} (\eu)         & $\mathbf{1.257 \pm 0.066}$ & $\mathbf{76.4}$ & 0.738 & 0.973 & 0.00064 & 0.0014 \\
M2 \method{} (\grpo)       & $1.253 \pm 0.069$ & $76.1$  & 0.735 & 0.973 & 0.00065 & 0.0014 \\
Oracle                     & $1.58 \pm 0.07$   & $100$   & --    & --    & --     & 0.0011 \\
\bottomrule
\end{tabular}
\end{table*}

\section{Results}

Table~\ref{tab:main} presents the main results on the card-fraud task, where the amount-scaled cost geometry leaves genuine decision headroom. We scope the study to this task deliberately: two other financial event tasks we initially considered---customer churn (D2) and loan default (D3)---turned out to have essentially \emph{no} decision headroom under any reasonable cost matrix, so post-hoc thresholding is already near-optimal and there is nothing for decision-focused fine-tuning to add. We summarize those observations in Appendix~\ref{app:d2d3}.

\paragraph{D1: Card-fraud screening (headline)}
\label{sec:results_d1}

On the high-frequency fraud task---24.4M transactions, 0.11\% prevalence, amount-scaled stakes---\method{} (\eu) achieves \textbf{1.257 utility per 1{,}000 transactions}, a $+0.084$ absolute gain over the STEP-CE + Bayes baseline ($1.173$), closing 76.4\% of the always-pass$\to$oracle gap (vs.\ 70.4\% for B3). The improvement is \emph{consistent}: all ten seeds are positive. A paired two-sided Wilcoxon signed-rank test gives $p = 0.0020$ ($n{=}10$), certifying the effect at $\alpha{=}0.05$ (see \S\ref{sec:significance}). The sampled variant (\grpo) matches \eu{} at 1.25, confirming the two objectives agree when the exact expectation is available.

The gain is not recalibration, and we can now bound that claim rather than assert it. Post-hoc calibrators are monotone maps of the predicted scores, so they cannot change the ranking---only where the amount-dependent threshold $t^*(m)$ bites. Isotonic regression is the most expressive monotone map available, which makes it an empirical ceiling on the whole class. Across all ten seeds, against a B3 of $1.173$, temperature scaling returns $+0.010$ ($8/10$ seeds, $p{=}0.16$), Platt $+0.026$ ($6/10$, $p{=}0.13$), and isotonic---the ceiling---$+0.031$ ($8/10$, $p{=}0.037$): only the nonparametric member separates from B3 at all. \method{} returns $+0.084$ on the same decision sets and beats \emph{every} calibrator on $10/10$ seeds (paired Wilcoxon $p{=}0.0020$ in each case; $+0.074$ vs.\ temperature scaling, $+0.058$ vs.\ Platt, $+0.052$ vs.\ isotonic). \textbf{The entire post-hoc calibration class accounts for at most $37\%$ of \method{}'s gain; the remaining $63\%$ requires changing the ranking, which no monotone recalibration can do.}

The mechanism is visible in the metrics, and it is the opposite of what a calibration story predicts. The calibrators are \emph{better} calibrated than \method{}---isotonic reaches ECE $0.8{\times}10^{-4}$ against \method{}'s $6.4{\times}10^{-4}$, a factor of eight---and still lose on utility. What \method{} has instead is a better ranking: PR-AUC $0.738$ versus $0.690$ for B3, temperature scaling and Platt alike (identical to three decimal places, as monotonicity requires). Calibration is not the binding constraint here; \emph{which events are ranked as risky} is. Isotonic even illustrates the cost of its own flexibility---by tying $1.8$M distinct scores into $\approx\!144$ steps it gives back $0.011$ of PR-AUC, the only one of the three that measurably coarsens the ranking. \method{} shifts the ranking toward high-stakes events because the per-decision gradient scales with the amount at stake (\S\ref{sec:linear}), at essentially unchanged ROC-AUC ($0.973$) and intervention rate ($0.14\%$).

The sharpest contrast is with the cost-sensitive \emph{training} heuristics. At matched Stage-2 compute, class-weighted CE (B5), focal loss (B6), and oversampling (B7) all \emph{fail} under $0.11\%$ prevalence: they raise PR-AUC but push the operating point to $1.6$--$31\%$ intervention and $28$--$316\times$ worse calibration, driving realized utility to $-0.76$, $-28.8$, and $+0.06$ respectively---at or below the always-pass floor. Reweighting the loss trades exactly the calibration that the Bayes decision rule depends on; the closed-form expected-utility objective, which reweights the \emph{gradient} by stake while anchoring calibration through KL, avoids this failure mode. Granting those heuristics the same KL anchor (B8) helps but does not rescue them (\S\ref{sec:anchor_vs_loss}). This is the clearest evidence for the method: it is not that \eu{} beats thresholding by a wide margin, but that the naive alternatives to thresholding are actively harmful here.

The classical tabular baselines (B1 logistic regression, B2 gradient-boosted trees) obtain \emph{negative} utility on their own harness: engineered features do not separate fraud sharply enough to overcome the false-positive cost of blocking legitimate transactions.

\paragraph{Is it the anchor or the objective?}
\label{sec:anchor_vs_loss}
B5--B7 as conventionally specified carry no KL anchor, while \method{} does. A comparison against them therefore changes two things at once, and the natural objection is that \method{}'s advantage is simply the anchor. B8 settles this. Adding \method{}'s exact KL term ($\beta{=}0.05$) to class-weighted CE, at matched Stage-2 budget on the same Stage-1 checkpoint, moves utility from $-0.76$ (B5) to $-0.34$ (B8): an improvement of $+0.42$, positive on $10/10$ seeds, paired Wilcoxon $p{=}0.0020$. The anchor is genuinely doing work---but less than its utility effect suggests. Measured against B3's ECE of $0.0003$, class weighting inflates calibration error by $0.0124$ and the anchor claws back only $0.0020$ of that ($16\%$); the intervention rate recovers comparably ($2.44\%\to1.99\%$ against B3's $0.12\%$, $19\%$ of the excess).

But it is not sufficient. Anchored class-weighted CE remains \emph{below the always-pass floor} ($-0.34$ vs.\ $0.19$), at $-39\%$ of the oracle gap. The remaining distance to \method{} (\eu) is $+1.59$ utility per 1{,}000 decisions, again $10/10$ seeds and $p{=}0.0020$. So of the total $+2.02$ separating B5 from M1, the anchor accounts for $21\%$ and the expected-utility loss form for $79\%$. Reweighting the \emph{loss} by class and then regularizing the damage is not a substitute for reweighting the \emph{gradient} by the stake actually at risk. We regard this, rather than the margin over B3, as the paper's central empirical claim.

\paragraph{How much anchoring? Sweeping $\beta$ and $\lambda$}
\label{sec:beta_sweep}
\S\ref{sec:anchor_vs_loss} shows the anchor is not what carries \method{}. The complementary question is whether \method{} is \emph{sensitive} to it---whether $(\beta,\lambda){=}(0.05,0.1)$ is a tuned magic number. We sweep
\begin{equation*}
\beta \in \{0, 0.01, 0.05, 0.2, 1.0\}\,, 
\end{equation*}
at $\lambda{=}0.1$, and $\lambda \in \{0, 0.1, 1.0\}$ at $\beta{=}0.05$, on seeds $\{41,42,43\}$, reporting every reference line on the same three seeds (Table~\ref{tab:beta}).

\emph{Utility is flat.} Every setting in the grid except $\beta{=}1.0$ lands in $1.28$--$1.32$, all comfortably above the seed-matched B3 value of $1.235$. Only $\beta{=}1.0$ degrades the result ($1.244$), as the anchor begins to dominate the utility signal and pulls the policy back toward the pretrained one. Neither coefficient needs tuning: the method's advantage is not purchased by hyperparameter search.

\emph{The defaults are not the utility optimum, and that is the point.} The best utility in the grid is $\beta{=}0.05,\lambda{=}0$ ($1.315$), marginally above the default $(0.05, 0.1)$ at $1.307$; and $\beta{=}0$---no anchor at all---still reaches $1.278$ and still beats B3. Read on utility alone, one would drop both anchors. But Table~\ref{tab:genfid} shows what that costs: at $\beta{=}0$ the generated fraud rate moves from $0.89{\times}$ reality to $1.58{\times}$ and the label marginal degrades $2.8$-fold. The anchors buy generative fidelity, not utility, and they buy it at a utility price of about $0.03$ per $1{,}000$ decisions---roughly a third of the headline gain, which we consider worth paying for a model that is supposed to remain a usable event simulator.

\emph{What $\beta$ actually controls} is calibration and intervention appetite, both monotone in it: ECE falls from $1.2{\times}10^{-3}$ at $\beta{=}0$ to $3.5{\times}10^{-4}$ at $\beta{=}1.0$, and the intervention rate from $0.22\%$ to $0.10\%$. $\beta$ is therefore best understood as the dial trading a little utility for calibration fidelity and operational conservatism---which is what an institution deploying this would want it to be.

Taken with B8, the attribution argument now closes from both sides: adding the anchor to a cost-sensitive loss does not rescue it (\S\ref{sec:anchor_vs_loss}), and removing the anchor from \method{} does not break it. The objective's form is doing the work.

\begin{table}[t]
\caption{Anchor-coefficient sweep on D1 (seeds $\{41,42,43\}$; all reference rows computed on the same three seeds). $(\beta,\lambda){=}(0.05,0.1)$ is the default used everywhere else in the paper. Utility is insensitive to both coefficients over the practical range; calibration and intervention rate improve monotonically with $\beta$. The utility-optimal settings are \emph{not} the defaults---see \S\ref{sec:genfid} for what the defaults buy instead.}
\label{tab:beta}
\begin{tabular}{@{}llccc@{}}
\toprule
$\beta$ & $\lambda$ & Utility/1k & ECE & Interv. \\
\midrule
$0$ (no anchor) & $0.1$ & $1.278 \pm 0.060$ & $0.00116$ & $0.00220$ \\
$0.01$          & $0.1$ & $1.297 \pm 0.080$ & $0.00091$ & $0.00186$ \\
$0.05$          & $0.1$ & $1.307 \pm 0.084$ & $0.00055$ & $0.00134$ \\
$0.2$           & $0.1$ & $1.280 \pm 0.080$ & $0.00040$ & $0.00109$ \\
$1.0$           & $0.1$ & $1.244 \pm 0.084$ & $0.00035$ & $0.00102$ \\
\cmidrule(l){1-5}
$0.05$          & $0$   & $\mathbf{1.315 \pm 0.079}$ & $0.00060$ & $0.00120$ \\
$0.05$          & $1.0$ & $1.276 \pm 0.088$ & $0.00044$ & $0.00155$ \\
\cmidrule(l){1-5}
\multicolumn{2}{@{}l}{B3 CE + Bayes} & $1.235 \pm 0.067$ & $0.00024$ & $0.00129$ \\
\bottomrule
\end{tabular}
\end{table}

\paragraph{Does the result depend on the severity scale?}
\label{sec:scale_robustness}
Stakes enter the cost matrix through a monotone severity proxy $m$ derived from the amount bucket (\S\ref{sec:setup}), and a reader may reasonably ask whether the conclusions are an artifact of that encoding. Because the evaluation harness persists per-decision $(\hat p, y, m)$, we can re-score every method under a different severity scale \emph{without retraining}: the learned policies are fixed, and only the Bayes threshold $t^*(m)$ and the utility accounting change. Table~\ref{tab:scales} reports all three scales.

The conclusions are invariant. The method ordering is identical; \eu{} beats B3 on $10/10$ seeds with $p{=}0.0020$ under every scale; all of B5--B8 remain at or below the always-pass floor under every scale. The dollar-proportional scale also lets us state the headline in money: at a top-bucket median ticket of $\$251$, \eu{} realizes $\$135.0$ per $1{,}000$ decisions against $\$123.2$ for B3---a gain of $\mathbf{+\$11.8}$ per $1{,}000$ screened transactions, against an oracle ceiling of $\$183.5$ and an always-pass floor of $-\$5.8$. Over the $2.4$M-decision test horizon of a single seed that is roughly $\$28$k of recovered value from a one-hour fine-tune.

\begin{table}[t]
\caption{Severity-scale robustness on D1 (utility per 1{,}000 decisions, mean over 10 seeds; \%gap in parentheses). \emph{As-run} is the scale the models were trained and originally scored under; \emph{rank} is the normalized bucket rank $(b{+}0.5)/32$; \emph{dollars} is proportional to the bucket median ticket. Policies are unchanged across columns---only $t^*(m)$ and the accounting differ. Every ordering, sign and significance level is preserved.}
\label{tab:scales}
\begin{tabular}{@{}lccc@{}}
\toprule
Method & As-run & Rank & Dollars \\
\midrule
Always-pass           & $0.19$  & $0.21$  & $-0.02$ \\
B3 STEP-CE + Bayes    & $1.17\;(70.4)$  & $1.15\;(69.2)$  & $0.49\;(68.2)$ \\
B5 Weighted-CE        & $-0.76\;(-69)$  & $-0.82\;(-78)$  & $-0.48\;(-61)$ \\
B6 Focal              & $-28.8$ & $-30.3$ & $-11.2$ \\
B7 Oversample         & $0.06\;(-10)$   & $0.01\;(-16)$   & $-0.08\;(-8)$ \\
B8 Weighted-CE + KL   & $-0.34\;(-39)$  & $-0.39\;(-46)$  & $-0.28\;(-35)$ \\
M1 \method{} (\eu)    & $\mathbf{1.26}\;(\mathbf{76.4})$ & $\mathbf{1.23}\;(\mathbf{75.4})$ & $\mathbf{0.54}\;(\mathbf{74.3})$ \\
M2 \method{} (\grpo)  & $1.25\;(76.1)$  & $1.23\;(75.1)$  & $0.53\;(73.9)$ \\
Oracle                & $1.58\;(100)$   & $1.56\;(100)$   & $0.73\;(100)$ \\
\midrule
M1 $-$ B3             & $+0.084$ & $+0.085$ & $+0.047$ \\
\quad seeds positive  & $10/10$  & $10/10$  & $10/10$ \\
\quad Wilcoxon $p$    & $0.0020$ & $0.0020$ & $0.0020$ \\
\bottomrule
\end{tabular}
\end{table}

\paragraph{Generative fidelity: do the anchors preserve the event model?}
\label{sec:genfid}
Unlike a discriminative classifier, STEP is a \emph{generative} model of the whole event row, so it can roll out plausible \emph{future} transactions given a history---a capability we now use to test a standing claim of the method: that the KL and LM anchors (\S\ref{sec:linear}) preserve the pretrained generative event model while the label policy is re-aligned. This is also an axis on which a purely discriminative fix cannot be evaluated at all.

\paragraph{Protocol.} On the D1 test users we collect fully-real windows of $10$ event rows, condition each model on the first $5$ rows, and autoregressively sample the remaining $5$ (column-constrained decoding, temperature $1.0$, $20$ samples per window, $3{,}000$ windows, seeds $\{41,42,43\}$). We compare the generated continuations to the \emph{real} held-out continuation of the same window along three axes: the total-variation (TV) distance between generated and real token marginals, averaged over the categorical columns (cat-TV) and reported separately for the fraud-label column (label-TV); the Kolmogorov--Smirnov distance between generated and real amount-severity distributions (amt-KS); and the ratio of the generated to the real fraud base rate over the horizon (fraud-rate ratio; $1.0$ is perfect). Every model scores the identical windows.

\paragraph{Result.} Table~\ref{tab:genfid} shows a clean separation that mirrors, and extends, the calibration story of \S\ref{sec:results_d1}. The anchored objectives (\eu, \grpo) reproduce the pretrained model's fraud-label marginal exactly (label-TV $4$--$5{\times}10^{-4}$, indistinguishable from CE) and keep the generated fraud base rate close to reality (ratio $0.87$--$0.89$ against CE's $0.64$, though the seed-to-seed spread on this statistic is wide enough that we do not claim \eu{} \emph{beats} CE here---only that it does not degrade), while \emph{improving} categorical fidelity (cat-TV $0.242$ vs.\ CE $0.284$) and amount fidelity (amt-KS $0.100$ vs.\ $0.197$)---the amount-weighted objective sharpens the stake-bearing columns. The cost-sensitive training heuristics do the opposite: they corrupt the generative event model, inflating the fraud-label TV by $26$--$400{\times}$ and \emph{over-generating} fraud by $9$--$136{\times}$ (focal is worst, hallucinating fraud at $136{\times}$ the real rate). The damage these heuristics do is thus not confined to the decision layer's calibration (Table~\ref{tab:main}); it reaches into the joint event distribution the model was pretrained to represent.

\paragraph{Is it the anchor that prevents this?} Only partly---and the same $2{\times}2$ that settled the utility question (\S\ref{sec:anchor_vs_loss}) settles this one, in the same direction. Giving class-weighted CE the KL anchor (B8) improves label-TV by just $17\%$ ($18.9 \to 15.7 \times 10^{-3}$) and the fraud-rate ratio by $16\%$ ($15.6 \to 13.1$): the anchored heuristic still over-generates fraud by an order of magnitude and remains $39{\times}$ worse than CE on the label marginal. Conversely, stripping the KL anchor out of \method{} ($\beta{=}0$) leaves cat-TV and amt-KS \emph{untouched} ($0.242$, $0.100$---identical to anchored \eu) but degrades the label marginal by $2.8{\times}$ ($0.40 \to 1.11 \times 10^{-3}$) and pushes the fraud rate from $0.89$ to $1.58$, i.e.\ from slightly under-generating to over-generating.

So the anchor is real but local: it governs exactly the one column the objective applies pressure to, and leaves the rest of the event model alone. What protects the other thirteen columns is the \emph{loss form}---an objective that reweights the gradient by stake rather than reweighting the data by class. This is the same $\approx$$20/80$ split between anchor and objective that we measured on utility, arrived at through an entirely independent measurement, and it is the reason we report a nonzero default $\beta$ even though \S\ref{sec:beta_sweep} shows utility is nearly flat in it.

\begin{table*}[t]
\caption{Generative fidelity on D1 (mean $\pm$ s.d.\ over seeds $\{41,42,43\}$; $3{,}000$ held-out windows, $20$ samples each). Each model conditions on $5$ real event rows and generates the next $5$; generated columns are compared to the real continuation. Lower is better for cat-TV, label-TV, amt-KS; the fraud-rate ratio is best at $1.0$. The middle block is the anchor $2{\times}2$: adding the anchor to a heuristic loss (B5$\to$B8) barely helps, while removing it from \method{} ($\beta{=}0$) costs a factor of $2.8$ in label-TV. Loss form dominates; the anchor is a second-order correction on top of the right loss.}
\label{tab:genfid}
\begin{tabular}{@{}lcccc@{}}
\toprule
Method & cat-TV $\downarrow$ & label-TV $\downarrow$ & amt-KS $\downarrow$ & fraud ratio \\
 & & ($\times10^{-3}$) & & ($\to 1$) \\
\midrule
B3 STEP-CE       & $0.284{\pm}0.003$ & $0.40{\pm}0.26$   & $0.197{\pm}0.066$ & $0.64{\pm}0.26$ \\
M1 \method{} (\eu)   & $\mathbf{0.242}{\pm}0.008$ & $\mathbf{0.40}{\pm}0.11$ & $\mathbf{0.100}{\pm}0.043$ & $\mathbf{0.89}{\pm}0.39$ \\
M2 \method{} (\grpo) & $0.242{\pm}0.008$ & $0.48{\pm}0.16$   & $0.100{\pm}0.043$ & $0.87{\pm}0.48$ \\
\cmidrule(l){1-5}
\multicolumn{5}{@{}l}{\emph{anchor controls (the $2{\times}2$)}}\\
\eu, $\beta{=}0$ (no anchor) & $0.242{\pm}0.010$ & $1.11{\pm}0.85$ & $0.100{\pm}0.053$ & $1.58{\pm}0.77$ \\
B5 Weighted-CE   & $0.258{\pm}0.004$ & $18.9{\pm}8.1$    & $0.103{\pm}0.025$ & $15.6{\pm}2.8$ \\
B8 Weighted-CE + KL & $0.258{\pm}0.004$ & $15.7{\pm}8.0$ & $0.103{\pm}0.030$ & $13.1{\pm}2.6$ \\
\cmidrule(l){1-5}
B6 Focal         & $0.257{\pm}0.004$ & $160.6{\pm}15.8$  & $0.100{\pm}0.033$ & $135.8{\pm}28.6$ \\
B7 Oversample    & $0.257{\pm}0.003$ & $10.4{\pm}3.4$    & $0.104{\pm}0.034$ & $9.2{\pm}0.5$ \\
\bottomrule
\end{tabular}
\end{table*}

\paragraph{Statistical significance}
\label{sec:significance}
For the primary comparison (M1 \eu{} vs.\ B3 CE+Bayes) we pre-registered a paired \emph{two-sided} Wilcoxon signed-rank test. The result is $\Delta{=}{+}0.084$, $W{=}0$, $p{=}0.0020$ ($n{=}10$), significant at $\alpha{=}0.05$ (one-sided $p{=}0.0010$, reported as secondary), with all ten seeds positive. The certified effect confirms that decision-focused fine-tuning yields a reliable utility gain where the amount-scaled cost geometry leaves room to improve decisions.

The same test applied to the recalibration baselines gives $\Delta{=}{+}0.074$ vs.\ B4a, $+0.058$ vs.\ B4b and $+0.052$ vs.\ B4c, each with $W{=}0$, $p{=}0.0020$ and $10/10$ seeds positive. Because all four arms are scored on the identical per-seed decision set, these are exactly paired. We note the converse for completeness: of the calibrators themselves, only isotonic separates from B3 at $\alpha{=}0.05$ ($+0.031$, $p{=}0.037$); temperature scaling ($+0.010$, $p{=}0.16$) and Platt ($+0.026$, $p{=}0.13$) do not. Each of the four secondary comparisons has $p{=}0.0020 < \alpha/4 = 0.0125$, so the conclusion survives a Bonferroni multiplicity adjustment.

\begin{figure}[t]
\centering
\includegraphics[width=\linewidth]{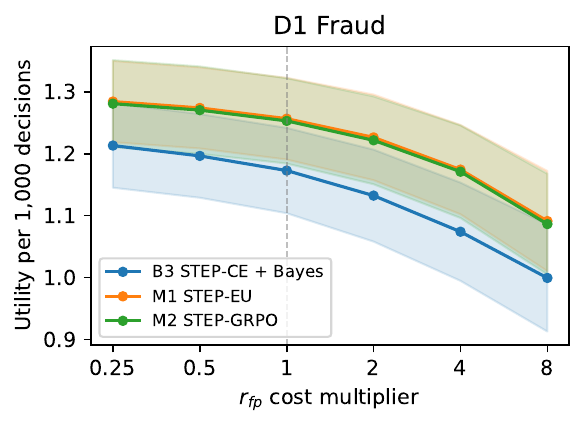}
\caption{Robustness to FP-cost misspecification. Utility (mean $\pm$ 1~s.d.\ across seeds) as $r_{fp}$ is scaled at evaluation time. \eu{} dominates B3 across the full range ($0.25$--$8{\times}$).}

\label{fig:robustness}
\end{figure}

\paragraph{Robustness to cost misspecification}
\label{sec:robustness_results}
A deployed cost matrix is always an approximation. Figure~\ref{fig:robustness} evaluates realized utility when the true $r_{fp}$ differs from the training-time value by a factor $\in\{0.25,0.5,1,2,4,8\}$; for each multiplier we recompute the Bayes-optimal threshold $t^*$ and re-score the test stream. \eu{} dominates B3 across the entire range: $+0.07$ at $0.25{\times}$ (1.28 vs.\ 1.21), $+0.08$ at $0.5{\times}$ (1.27 vs.\ 1.20), $+0.08$ at nominal (1.26 vs.\ 1.17), $+0.09$ at $2{\times}$ (1.23 vs.\ 1.13), $+0.10$ at $4{\times}$ (1.17 vs.\ 1.07), and $+0.09$ at $8{\times}$ (1.09 vs.\ 1.00), positive on $10/10$ seeds at every multiplier except $8{\times}$ ($9/10$). Both curves decline smoothly as the FP cost rises, as they should. The \eu{} advantage is thus not an artifact of one FP-cost choice. (At $0.25{\times}$ the margin $C{+}G = (r_{tn}{-}r_{fp}) + (r_{tp}{-}r_{fn})$ is negative for low-stake events, where blocking a fraud is worth less than the review it costs; the Bayes rule there is never to intervene, and we apply it in the division-free form $p\,(C{+}G) \ge C$, which is valid for either sign.)


\paragraph*{Cost-curve analysis (D1)}
\label{sec:cost_curve}

\begin{figure}[t]
\centering
\includegraphics[width=\linewidth]{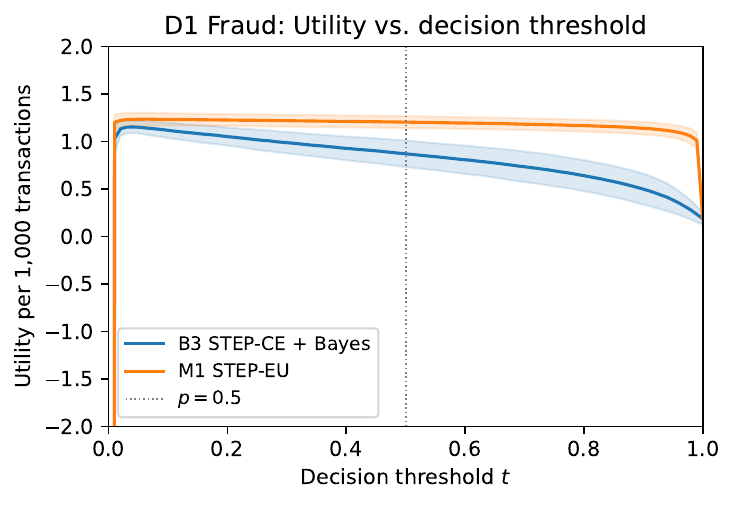}
\caption{D1: Utility per 1{,}000 transactions as a function of the decision threshold $t$. Shaded bands show $\pm 1$ s.d.\ across 10 seeds.}
\label{fig:cost_curve}
\end{figure}

Figure~\ref{fig:cost_curve} plots realized utility as a function of the global decision threshold $t$ for B3 and \eu{} on D1. The \eu{}-fine-tuned model attains a higher peak (1.24 at $t\!\approx\!0.07$ vs.\ 1.15 at $t\!\approx\!0.04$ for B3; these best single global thresholds sit just below the amount-dependent Bayes utilities of Table~\ref{tab:main} because a global threshold cannot exploit the per-event stake) and stays at or above the CE curve across the low-threshold region where the Bayes-optimal operating point lies, consistent with the modest PR-AUC improvement rather than a shift of operating point alone.

\section{Limitations and Ethical Considerations}
\label{sec:limitations}
\paragraph{No temporal-drift test.} Our split is by client, not by time (\S\ref{sec:protocol}), so the results establish generalization to unseen clients and say nothing about robustness to concept drift---precisely the regime in which a fraud model degrades in production. A rolling-origin evaluation, in which Stage 2 is refit on an expanding window and scored on the next period, is the natural next experiment; we expect the decision-focused stage to help more there, not less, since it is cheap enough to rerun each period, but we have not shown it.

\paragraph{One dataset, synthetic.} The study is scoped to a single fraud-screening task on a simulated corpus. Absolute dollar magnitudes should not be transported to production, and we do not claim the utility gains transfer to decision problems whose cost geometry removes decision headroom, where post-hoc thresholding is already optimal (Appendix~\ref{app:d2d3}). Validation on a real institutional event stream is the main outstanding question.

\paragraph{Refunds are mis-staked.} Quantile bucketing of a signed \texttt{Amount} places the $5.10\%$ of rows that are refunds or reversals---including $1{,}125$ fraud rows, $8.20\%$ of fraud by $|\$|$---in the lowest-stake bins (\S\ref{sec:setup}). The severity proxy therefore systematically under-weights refund fraud. We retain the convention for comparability with prior work on this corpus, but it makes our reported gains conservative, and $|\texttt{Amount}|$ is the better input in any deployment.

\paragraph{Stylized rewards.} Real institutions face review-capacity constraints, adversarial adaptation by fraudsters, and delayed or censored outcomes. Our one-step bandit ignores the effect of interventions on future behavior (blocking a card changes the subsequent stream); the \grpo{} machinery is the designated extension to trajectory-level rewards.

\paragraph{Fairness.} Cost matrices encode policy choices with fairness implications, friction costs may not be uniform across customer groups; we report intervention rates and urge subgroup audits before deployment. Only a public research dataset is used.

\section*{Conclusion}
A pretrained autoregressive tabular transformer already contains the decision layer a financial institution needs; one cheap, closed-form fine-tuning stage aligns it with the utility at stake. The objective is example-dependent cost-sensitive learning---linear in the predicted probability, sampling-free for matrix rewards, and anchored to the pretrained model. On a high-imbalance card-fraud task with genuine decision headroom, the alignment delivers a small but consistent utility gain over post-hoc Bayes thresholding, and cleanly outperforms cost-sensitive training heuristics that collapse under imbalance; the KL/LM anchors additionally preserve the pretrained generative event model that those heuristics corrupt. Evaluated the way it will be used---in realized utility---the method is a safe, cheap addition when the decision problem has room to improve.

\begin{acks}
The views and conclusions expressed in this work are those of the authors alone and do not necessarily reflect the positions of their affiliated institutions, employers, or organizations. This manuscript is intended solely for academic and research purposes and does not represent an endorsement, official position, or policy of any entity. No proprietary, customer, or internal data were used in the experiments described. Code is available from the authors upon reasonable request, subject to any required approvals.
\end{acks}

\bibliographystyle{ACM-Reference-Format}
\bibliography{references}

\appendix
\section{Exactness of the EU gradient}
For a one-step bandit with full counterfactual feedback,
\begin{equation*}
\nabla_\theta\, \mathbb{E}_{a\sim\pi_\theta}[R(a)] = \sum_a R(a)\,\nabla_\theta \pi_\theta(a) = \mathbb{E}_{a\sim\pi_\theta}\!\big[R(a)\,\nabla_\theta\log\pi_\theta(a)\big],
\end{equation*}
and the middle expression is computable directly since $R(a)$ is known for all $a$; hence \eu{} equals the REINFORCE expectation with zero estimator variance. The \grpo{} variant we report uses a leave-one-out group baseline; the subtracted mean is a control variate that leaves the expected gradient unchanged, so $\mathbb{E}[\nabla_{\grpo}] = \nabla_{\eu}$ exactly (a baseline shifts variance, not direction). The alternative std-normalized advantage additionally divides by the group reward standard deviation, making $\mathbb{E}[\nabla]$ a per-decision positive rescaling of $\nabla_{\eu}$ where that std is well-defined---but it is undefined for homogeneous groups, which can collapse learning under peaked, imbalanced policies.

\section{Reproducibility}
Code is available upon request, subject to institutional review and approval. Namely: RL objectives, trainer, configs (including the cost matrices of Table~\ref{tab:costs}), evaluation scripts producing every number in this paper (including the generative-fidelity rollout of \S\ref{sec:genfid}, \texttt{generate\_trajectory\_eval.py}), and unit tests.

\paragraph{Stage-1 hyperparameters.} GPT-2-small decoder (12 layers, 12 heads, 768 hidden); batch size 32, sequence length 10 rows, gradient accumulation 8 steps, lr $10^{-5}$ with cosine schedule and 1{,}000 warmup steps, 2 epochs. Client-level split, 1{,}600/200/200 users (\S\ref{sec:protocol}). Checkpoint selection by validation AUC.

\paragraph{Stage-2 hyperparameters.} AdamW, lr $10^{-6}$, 3 epochs, $\beta = 0.05$ (KL coefficient), $\lambda = 0.1$ (LM coefficient), reward scale $s = 1.0$, GRPO group size $G = 8$. Checkpoint selection by validation utility (Bayes threshold).

\paragraph{Baseline hyperparameters (B5--B8).} Class-weighted CE (B5) and focal loss (B6) use an inverse-frequency positive weight $n_{\mathrm{neg}}/n_{\mathrm{pos}} \approx 900$ for D1; focal loss uses focusing parameter $\gamma = 2.0$~\cite{lin2017focal}. Oversampling (B7) balances the two classes exactly (computed in closed form as inverse in-batch class-frequency weighting). All of B5--B7 receive the same auxiliary-CE coefficient $\lambda = 0.1$ and Stage-2 budget as \eu/\grpo{} but no KL anchor. B8 is B5 with \method{}'s KL-to-reference term at the identical coefficient $\beta = 0.05$ and an otherwise identical configuration, so B8 $-$ B5 isolates the anchor and M1 $-$ B8 isolates the loss form (\S\ref{sec:anchor_vs_loss}).

\paragraph{Seeds.} Seeds $\{41, 42, \ldots, 50\}$ (ten seeds).

\paragraph{Compute.} All runs on a single A100-80GB. Stage 1: $\sim$5 hours per seed. Stage 2: to keep fine-tuning a small fraction of pretraining, \eu/\grpo{} train for 3 epochs over a fixed $10^5$-sequence subsample ($\sim$1M decisions), $\sim$1 hour per seed (roughly 20\% of Stage 1). Every cost-sensitive baseline (B5--B8) is granted the identical Stage-2 budget on the same Stage-1 checkpoint, so all comparisons are matched-compute continuations.

\paragraph{Software.} PyTorch 2.7, Transformers 5.12, Hydra 1.3. No distributed training; single GPU per run.

\section{Excluded tasks: churn and loan default}
\label{app:d2d3}
Beyond card fraud, we also evaluated the method on two additional public financial event tasks: \textbf{D2 --- customer churn} on the Rosbank transaction dataset (predict-and-retain: offer a costly retention incentive to a customer about to leave) and \textbf{D3 --- loan default} on the PKDD'99 (Berka) Czech banking dataset (predict-and-intervene: act on a loan likely to default). We ran the full pipeline of Sections~\ref{sec:setup}--\ref{sec:results_d1} on both---STEP-CE with Bayes thresholding, the cost-sensitive heuristics, and \method{} (\eu/\grpo)---but do not feature them in the main paper. This appendix records why.

The reason is a lack of \emph{decision headroom}. Unlike fraud (D1), where positives are vanishingly rare ($\approx0.1\%$) and the amount-scaled cost matrix places the Bayes-optimal threshold in a regime where the model's ranking genuinely changes which transactions are blocked, D2 and D3 have moderate base rates and (once the retention/recovery economics are written down) cost matrices whose Bayes-optimal action is to intervene on almost \emph{everyone}: the induced intervention rate sits at essentially $1.0$ (Table~\ref{tab:d2d3}). In that regime the decision is dominated by the cost geometry rather than by the model's probabilities, so post-hoc thresholding of the pretrained STEP-CE model already captures the great majority of the available value---roughly $86$--$92\%$ of the always-pass$\to$oracle gap on these two tasks---and leaves almost nothing for a decision-focused stage to recover. Consistent with this, \method{} (\eu{} and \grpo) matched the STEP-CE + Bayes utility to within noise on both tasks (neither improving nor harming it), exactly as the theory predicts when the Bayes rule is already pinned to a corner of the action simplex. The cost-sensitive training heuristics remained harmful (as on fraud), but since the honest baseline was already near-optimal, the tasks could not distinguish our method from simple thresholding and added no evidence for the central claim.

\begin{table}[t]
\caption{Summary of the excluded tasks (mean over seeds; D2 uses 5 seeds, D3 uses 10). Utility is per 1{,}000 decisions on each task's \emph{own} reward scale and is \emph{not} comparable across tasks or to D1; \%gap is the share of the always-pass$\to$oracle gap closed by the Bayes rule. On both tasks the Bayes-optimal intervention rate is $\approx 1.0$, so STEP-CE + Bayes already closes most of the gap and \method{} (\eu/\grpo) ties it to within noise---i.e.\ no decision headroom.}
\label{tab:d2d3}
\begin{tabular}{@{}lccc@{}}
\toprule
Method & Utility/1k & \%gap & Interv.\ rate \\
\midrule
\multicolumn{4}{@{}l}{\emph{D2 --- churn (Rosbank)}} \\
STEP-CE + Bayes    & $49.4$ & $91.9$ & 1.00 \\
\method{} (\eu)    & $49.4$ & $91.9$ & 1.00 \\
\method{} (\grpo)  & $49.4$ & $91.9$ & 1.00 \\
\cmidrule(l){1-4}
\multicolumn{4}{@{}l}{\emph{D3 --- loan default (PKDD'99)}} \\
STEP-CE + Bayes    & $23.4$ & $86.0$ & 1.00 \\
\method{} (\eu)    & $23.2$ & $85.9$ & 0.99 \\
\method{} (\grpo)  & $22.9$ & $85.7$ & 0.98 \\
\bottomrule
\end{tabular}
\end{table}

We therefore focus the paper on the card-fraud task (D1), whose extreme imbalance and amount-scaled stakes leave real room to improve decisions and thus make it the setting where decision-focused fine-tuning can be meaningfully evaluated. We view D2 and D3 not as negative results for the method but as cases outside its intended operating regime: when post-hoc thresholding is already optimal, no fine-tuning objective---ours or otherwise---should be expected to help.

\end{document}